%% file: emnlp2023.tex
\pdfoutput=1

\documentclass[11pt]{article}

\usepackage[]{EMNLP2023}

\usepackage{times}
\usepackage{latexsym}

\usepackage[T1]{fontenc}

\usepackage[utf8]{inputenc}

\usepackage{microtype}

\usepackage{inconsolata}

\usepackage{xcolor}
\PassOptionsToPackage{dvipsnames}{xcolor}
\usepackage{multirow}
\usepackage{multicol}
\usepackage{graphicx}
\usepackage{tabu}
\usepackage{amsmath}
\usepackage{booktabs}
\usepackage{hyperref}

\usepackage{subcaption}
\usepackage{caption}
\usepackage[normalem]{ulem}
\usepackage{soul}
\usepackage[shortlabels]{enumitem}
\usepackage{array}
\usepackage{pgffor}
\usepackage{multirow}
\usepackage{cancel}
\usepackage[normalem]{ulem}
\newcolumntype{P}[1]{>{\centering\arraybackslash}p{#1}}

\definecolor{light_green}{rgb}{0.56, 0.93, 0.56}
\usepackage[most]{tcolorbox}
\tcbset{on line, 
        boxsep=1pt, left=0pt,right=0pt,top=0pt,bottom=0pt,
        colframe=white,
        colback=light_green,  
        highlight math style={enhanced}
        }

\definecolor{light_red}{rgb}{1.0, 0.6, 0.6}

\newcommand{\dataset}{\textsc{LongBoX}}

\title{LongBoX: Evaluating Transformers on Long-Sequence Clinical Tasks}

\author{Mihir Parmar$^1$ \quad \quad Aakanksha Naik$^2$ \quad \quad Himanshu Gupta$^1$\\\textbf{Disha Agrawal}$^1$ \quad \quad \textbf{Chitta Baral}$^1$ \\ \\
$^1$Arizona State University, USA \\
$^2$Allen Institute of AI, USA
}

\begin{document}
\maketitle

\input{0_abstract}
\input{1_introduction}
\input{2_lilo}
\input{3_model_performance}
\input{4_related_work}
\input{5_conclusion}

\bibliography{anthology,custom}
\bibliographystyle{acl_natbib}


\end{document}

%% file: 0_abstract.tex
\begin{abstract}

Many large language models (LLMs) for medicine have largely been evaluated on short texts, and their ability to handle longer sequences such as a complete electronic health record (EHR) has not been systematically explored. Assessing these models on long sequences is crucial since prior work in the general domain has demonstrated performance degradation of LLMs on longer texts. Motivated by this, we introduce \dataset{}, a collection of seven medical datasets in text-to-text format, designed to investigate model performance on long sequences. Preliminary experiments reveal that both medical LLMs (e.g., BioGPT) and strong general domain LLMs (e.g., FLAN-T5) struggle on this benchmark. We further evaluate two techniques designed for long-sequence handling: (i) local-global attention, and (ii) Fusion-in-Decoder (FiD). Our results demonstrate mixed results with long-sequence handling - while scores on some datasets increase, there is substantial room for improvement. We hope that \dataset{} facilitates the development of more effective long-sequence techniques for the medical domain\footnote{\url{https://github.com/Mihir3009/LongBoX}}.


\end{abstract}

%% file: 1_introduction.tex
\section{Introduction}

In recent years, the exponential increase in machine-readable text in the medical domain such as electronic health records (EHRs) has sparked a growing interest in the development of pretrained medical language models \cite{lewis-etal-2020-pretrained}. Over the years, many large language models (LLMs) have been developed in these domains such as BioGPT \cite{luo2022biogpt}, BioMedLM \cite{venigalla2022biomedlm}, GatorTRONGPT \citep{peng2023study} and MedPaLM \citep{singhal2022large}. These LLMs have been evaluated on a wide range of medical tasks, but most tasks have only involved short texts. Many real-world medical tasks on the other hand require models to make predictions from longer texts, such as a summary from a patient visit or a series of EHRs for a patient, hence evaluating performance on longer texts is crucial. While this problem has been tackled in the general domain \cite{shaham-etal-2022-scrolls, tay2021long}, model ability to handle long sequences in the clinical domain is under-explored.

To tackle this, we propose \dataset, a collection of seven carefully-curated clinical datasets, which can measure performance of models on long sequences, converted to a unified text-to-text format. \dataset{} incorporates three task types: text classification, relation extraction and multi-label classification, and several types of clinical inputs such as discharge summaries and longitudinal records. Most importantly, for all datasets, input texts typically contains thousands of words. 


\input{tables/data_desc}


We first benchmark the performance of widely used high-performing LLMs on \dataset{} from general domain: LLaMA-2 \cite{touvron2023llama}, GPT-Neo \cite{black-etal-2022-gpt}, FLAN-T5 \cite{chung2022scaling} and from medical domain: SciFive \cite{phan2021scifive}, In-BoXBART \cite{parmar-etal-2022-boxbart}, Clinical-T5 \cite{lu-etal-2022-clinicalt5}, BioGPT \cite{luo2022biogpt}, and BioMedLM \cite{venigalla2022biomedlm}. Our results reveal that these models struggle on all datasets from \dataset{} achieving an average score of $\sim 52\%$. Next we evaluate two long sequence techniques that have shown promise in the general domain: (i) local-global attention (e.g., LongT5 \cite{guo-etal-2022-longt5}), and (ii) Fusion-in-Decoder (FiD) \cite{izacard-grave-2021-leveraging} (w/ SciFive and Clinical-T5). These methods achieve mixed results on \dataset{}, further highlighting the need for our benchmark, which we hope facilitates the development of better long sequence handling techniques for medical text. 

%% file: tables/data_desc.tex
\begin{table*}
\centering
\footnotesize
\resizebox{0.85\linewidth}{!}{
\begin{tabular}{c|c|ccc|c|c}
\toprule
\multirow{2}{*}{\textbf{Dataset}} & \multirow{2}{*}{\textbf{\begin{tabular}[c]{@{}c@{}}Document\\Types \end{tabular}}}  & \multicolumn{3}{c|}{\textbf{\# of Samples}}     &  \multirow{2}{*}{\textbf{\begin{tabular}[c]{@{}c@{}} Avg. \\ Tokens \end{tabular}}} & \multirow{2}{*}{\textbf{\begin{tabular}[c]{@{}c@{}} Max. \\ Tokens \end{tabular}}}  \\ \cmidrule{3-5}
                                  &                                                           & \multicolumn{1}{c|}{\textbf{Train}} & \multicolumn{1}{c|}{\textbf{Val}} & \multicolumn{1}{c|}{\textbf{Test}} &   &                                               \\ \midrule
Smoking 2006                      & DS                                                & \multicolumn{1}{c|}{358}            & \multicolumn{1}{c|}{40}           & 104           & 1187.98 & 3794                                             \\
Obesity 2008                      & DS                                             & \multicolumn{1}{c|}{11552}          & \multicolumn{1}{c|}{805}          & 7239          & 1903.47 & 4477                                               \\
Assertions 2010                   & DS, PR                                            & \multicolumn{1}{c|}{7073}           & \multicolumn{1}{c|}{1259}         & 11013             & 2169.40 & 5737                                                \\
Temporal RE                       & DS                                            & \multicolumn{1}{c|}{31513}          & \multicolumn{1}{c|}{2554}         & 22643         &  1154.68 & 2775                                               \\
RFHD 2014                         & LR                                            & \multicolumn{1}{c|}{4243}           & \multicolumn{1}{c|}{280}          & 2516          & 1152.14 & 4618                                               \\
Cohort Selection                  & LR                                              & \multicolumn{1}{c|}{2626}           & \multicolumn{1}{c|}{1118}         & 1118          & 6941.14 & 25608                                               \\
ADE 2018                          & DS                                            & \multicolumn{1}{c|}{36348}          & \multicolumn{1}{c|}{2346}         & 20593         &  4274.43 & 11550                                                \\ \bottomrule
\end{tabular}
}
\caption{An overview of document types used to create the dataset, along with a statistical analysis of each dataset. DS: Discharge Summaries, PR: Progress Reports, LR: Longitudinal Records}
\label{tab:stat_dataset}
\end{table*}


%% file: 2_lilo.tex
\section{LongBoX}
\label{sec:longbox}

\dataset{} consists of seven clinical datasets carefully curated from \textit{n2c2 NLP Research Datasets} collection\footnote{\url{https://portal.dbmi.hms.harvard.edu/projects/n2c2-nlp/}}:(1) Smoking Challenge Data 2006 \cite{uzuner2008identifying}, (2) Obesity Challenge Data 2008 \cite{uzuner2009recognizing}, (3) Assertions Challenge Data 2010 \cite{uzuner20112010}, (4) Temporal Relations 2012 \cite{sun2013evaluating}, (5) Heart Disease 2014 \cite{kumar2015creation}, (6) Cohort Selection 2018 \cite{stubbs2019cohort}, and (7) Adverse Drug Events (ADE) 2018 \cite{henry20202018}. Table \ref{tab:stat_dataset} presents details about the type of input text, dataset splits, and token length statistics for each dataset.

\subsection{Benchmark Details}

We provide a comprehensive overview of all datasets in \dataset{}, along with descriptions of diverse document types that were annotated to create these datasets.

\subsubsection{Document Types}
\paragraph{Discharge Summaries} are clinical notes containing details about why a person was admitted, diagnosis, medical regimen and response to their diagnosis, medical condition at discharge time, and after discharge care such as medications to continue at home \cite{kind2008documentation}. These summaries are in long text format but often not organized.

\paragraph{Progress Reports} are clinical documents that form the basis of the next plan of treatment. They consist of assessment, diagnosis, planning, intervention, and evaluation sections.

\paragraph{Longitudinal Records} are clinical documents that aggregate information from various sources in the health care system.

\subsubsection{Dataset Overview}
\paragraph{Smoking 2006} \cite{uzuner2008identifying}: Given discharge summaries for patients, the task is to categorize the smoking status of a patient into: (1) Past Smoker, (2) Current Smoker, (3) Smoker, (4) Non-Smoker, and (5) Unknown. This dataset was released as part of the n2c2 challenge in 2006. 

\paragraph{Obesity 2008} \cite{uzuner2009recognizing}: Based on discharge summaries, the task is to determine the presence of 15 different diseases such as asthma, and diabetes, which are potential indicators of obesity. The goal here is to categorize the presence of disease into: (1) Present, (2) Absent, (3) Questionable, and (4) Unmentioned. This dataset was released as part of the n2c2 challenge in 2008.

\paragraph{Assertions 2010} \cite{uzuner20112010}: Given discharge summaries as well as progress reports of patients, the task is to classify the occurrence of a concept into 6 categories: (1) Present, (2) Absent, (3) Hypothetical, (4) Possible, (5) Associated with someone else, and (6) Conditional. The concept can be medical problems, treatments, and tests. This dataset was released as part of the n2c2 challenge in 2010.


\paragraph{Temporal Relations 2012} \cite{sun2013evaluating}: The dataset consists of discharge summaries. Given a clinically significant event and time entity, the task is to find the type of relationship between them - BEFORE (event happens before given temporal expression), AFTER (event happens after given temporal expression), SIMULTANEOUS (event happens on given temporal expression), OVERLAP (event overlaps with temporal expression), BEGUN\_BY (event started on given temporal expression), ENDED\_BY (event ended on given temporal expression), DURING (event happens during given temporal expression), and BEFORE\_OVERLAP (event started before and lasts during given temporal expression). This dataset was released as part of the n2c2 challenge in 2012.

\paragraph{Heart Disease 2014} \cite{kumar2015creation}: This dataset consists of longitudinal medical records. The task here is to find indicators of a given condition in the text and classify them into "Present" and "Not present". For instance, the indicator for Diabetes can be different aspects such as the patient mentioning having diabetes, high glucose, and high HBA1c levels. This dataset was released as part of the n2c2 challenge in 2014.

\paragraph{Cohort Selection} \cite{stubbs2019cohort}: In this dataset, the goal is to classify whether a patient meets or does not meet specific criteria for participation in clinical trials. Clinical trials have certain criteria for including a patient in the trial group. The dataset includes 13 defined criteria such as MAJOR-DIABETES (Major diabetes-related complication), ALCOHOL-ABUSE (Current alcohol use over weekly recommended limits), and ENGLISH (Patient must speak English). This dataset was released as part of the n2c2 challenge in 2018.


\paragraph{ADE 2018} \cite{henry20202018}: Given discharge summaries, the task here is to classify the relationship between a drug and another related entity such as Strength-Drug (e.g., 20mg), Dosage-Drug (e.g., 1 tab per day), Duration-Drug (e.g., 5-day course), Frequency-Drug (e.g., every 4-6 hrs), Form-Drug (e.g., tablet, capsule), Route-Drug (e.g., intraperitoneal, IM), Reason-Drug (reason/disease for which the medication is prescribed), and ADE-Drug (side effect caused by the drug). This was another dataset released as part of the n2c2 challenge in 2018.

\subsection{Qualitative Analysis}

\begin{figure}
    \centering
    \includegraphics[width=\linewidth]{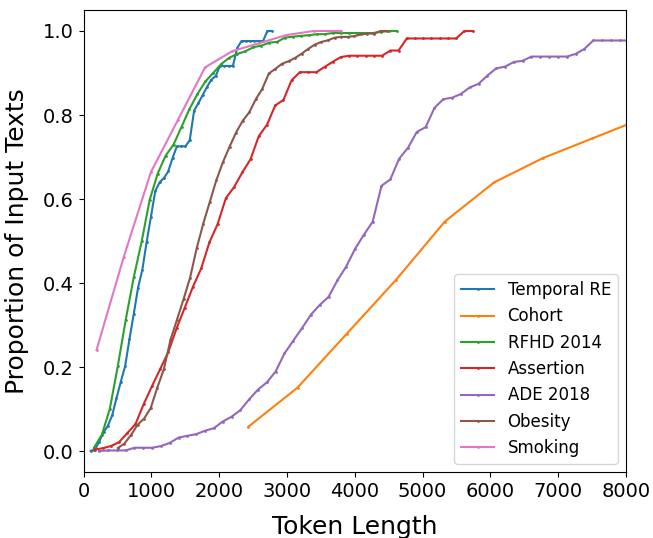}
    \caption{Cumulative distributions of input token lengths for all \dataset{} datasets.}
    \label{fig:hist}
\end{figure}

\paragraph{Length Analysis}
To analyze input lengths for all datasets, we first tokenize their test sets using the In-BoXBART\footnote{\url{https://huggingface.co/cogint/in-boxbart}} tokenizer. Table \ref{tab:stat_dataset} presents the average and maximum input token lengths for each dataset, which range from 1152-6941 and 2775-25608 respectively. Additionally, Figure \ref{fig:hist} displays cumulative distributions of input token lengths for each dataset (cut off at $8k$ for visibility) - given a token length $x$, the Y-axis indicates the proportion of inputs in the test set with token length $\leq x$. Though input token length shows considerable variation across datasets, it is clear that the maximum token limit for most LLMs (1024) is within $40^{th}$ percentile range for most datasets, and most of the instances within each dataset exceed $3k$ tokens.


\input{tables/baseline_results}

\paragraph{Comparing text lengths under regular and clinical tokenization (RoBERTa \textit{vs.} GatorTron)}

\begin{figure}
    \centering
    \includegraphics[width=\linewidth]{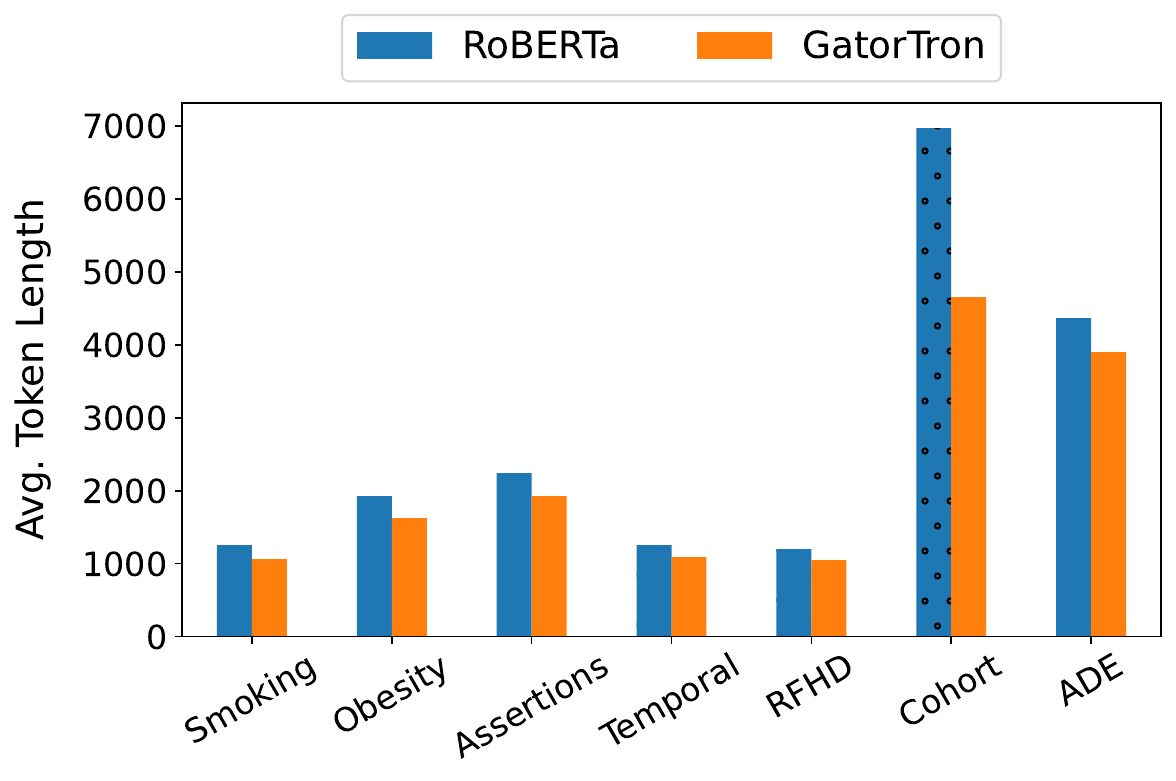}
    \caption{Average token length comparison between GatorTron and RoBERTa for all \dataset{} datasets.}
    \label{fig:token_length}
\end{figure}

To assess whether clinical tokenizers are able to significantly reduce text lengths over general domain tokenizers, we conduct a comparative analysis of text tokenized by GatorTron (a clinical model, with clinically-tailored tokenization) and RoBERTa (a general domain model). We tokenize the test sets of all datasets using both GatorTron and RoBERTa tokenizers. In Figure \ref{fig:token_length}, the average token lengths (on the $Y$-axis) for the test set of each dataset (on the $X$-axis) from \dataset{} are presented. From Figure \ref{fig:token_length}, it is evident that the clinical tokenizer generates shorter token lengths compared to the general domain tokenizer, though these differences are sometimes small. Notably, the difference between the average token lengths for clinical \textit{vs.} general tokenizers becomes larger as the input length increases, particularly observed in the cohort selection dataset.

%% file: tables/baseline_results.tex

\begin{table*}
\centering
\footnotesize
\resizebox{\linewidth}{!}{
\begin{tabular}{c|ccc|ccccc}
\toprule
                                     & \multicolumn{4}{c|}{$Enc. + Dec.$ Models} & \multicolumn{4}{c}{$Dec.$ Models}                                                              \\ \cmidrule{2-9} 
\multirow{-2}{*}{\textbf{Dataset}}            &  \multicolumn{1}{c|}{\textbf{FLAN-T5}}   & \multicolumn{1}{c|}{\textbf{In-BoXBART}} & \multicolumn{1}{c|}{\textbf{SciFive}} & \multicolumn{1}{c|}{\textbf{Clinical-T5}} & \multicolumn{1}{c|}{\textbf{GPT-Neo}}   & \multicolumn{1}{c|}{\textbf{BioGPT}} & \multicolumn{1}{c|}{\textbf{BioMedLM}} & \textbf{LLaMA-2} \\ \midrule

Smoking 2006                         &  \multicolumn{1}{c|}{55.77}         & \multicolumn{1}{c|}{58.65}          & \multicolumn{1}{c|}{60.58}  &   \multicolumn{1}{c|}{\textbf{64.41}}   & \multicolumn{1}{c|}{58.65}         & \multicolumn{1}{c|}{\textbf{59.62}}      & \multicolumn{1}{c|}{50.58} &    38.46     \\ 

Obesity 2008               &  \multicolumn{1}{c|}{68.28}         & \multicolumn{1}{c|}{\textbf{71.86}}          & \multicolumn{1}{c|}{\textbf{71.86}}   &  \multicolumn{1}{c|}{70.55}  & \multicolumn{1}{c|}{73.08}        & \multicolumn{1}{c|}{71.86}      & \multicolumn{1}{c|}{71.75} &    \textbf{84.78}     \\


Assertions 2010     &  \multicolumn{1}{c|}{\textbf{68.86}}         & \multicolumn{1}{c|}{67.95}          & \multicolumn{1}{c|}{67.83}   &  \multicolumn{1}{c|}{68.17}   & \multicolumn{1}{c|}{70.87}         & \multicolumn{1}{c|}{67.63}      & \multicolumn{1}{c|}{66.01} &   \textbf{73.09}      \\

Temporal RE                   &  \multicolumn{1}{c|}{\textbf{56.53}}         & \multicolumn{1}{c|}{56.36}          & \multicolumn{1}{c|}{56.03}    &  \multicolumn{1}{c|}{56.27}  & \multicolumn{1}{c|}{46.37}         & \multicolumn{1}{c|}{48.54}      & \multicolumn{1}{c|}{48.96}  &   \textbf{54.22}    \\

RFHD 2014                            &  \multicolumn{1}{c|}{64.99}         & \multicolumn{1}{c|}{\textbf{66.64}}          & \multicolumn{1}{c|}{64.58}  &  \multicolumn{1}{c|}{65.57}   & \multicolumn{1}{c|}{34.13}         & \multicolumn{1}{c|}{2.9}      & \multicolumn{1}{c|}{33.98} &    \textbf{74.76}     \\

Cohort Selection                     &  \multicolumn{1}{c|}{45.53}         & \multicolumn{1}{c|}{\textbf{47.67}}          & \multicolumn{1}{c|}{41.05}   & \multicolumn{1}{c|}{47.41}   & \multicolumn{1}{c|}{\textbf{55.90}}         & \multicolumn{1}{c|}{51.52}      & \multicolumn{1}{c|}{46.60} &    47.41    \\

ADE 2018                             &  \multicolumn{1}{c|}{19.07}         & \multicolumn{1}{c|}{17.62}          & \multicolumn{1}{c|}{\textbf{19.22}}  &   \multicolumn{1}{c|}{18.56}  & \multicolumn{1}{c|}{21.95}         & \multicolumn{1}{c|}{17.46}      & \multicolumn{1}{c|}{17.29} &   \textbf{23.97}      \\ \bottomrule

\end{tabular}
}
\caption{Performance of $Enc. + Dec.$ and $Dec.$ models on \dataset. All results are presented in \%.}
\label{tab:baseline_results}
\end{table*}

%% file: 3_model_performance.tex
\section{Experiments and Results}
\label{sec:model_learning}

\subsection{Experimental Setup}
\paragraph{Models}
We benchmark eight models from two architecture families: (i) four Encoder ($Enc.$) + Decoder ($Dec.$) models (FLAN-T5-Large from general domain; SciFive-Large, In-BoxBART, and Clinical-T5-Large from medical domain), and (ii) four Decoder ($Dec.$) only models (LLaMA-2-7B and GPT-Neo-1.3B from general domain; BioGPT-1.5B and BioMedLM-2.7B from medical domain). In addition, we evaluate two long sequence models. The first one is LongT5-Large, which enables a T5 encoder \cite{raffel2020exploring} to more efficiently handle long sequences by leveraging local-global attention sparsity patterns. The second is Fusion-in-Decoder (FiD), which breaks each input into smaller chunks, encodes them using an encoder-decoder model and then fuses encoded chunks in the decoder while generating output. We experiment with both SciFive and Clinical-T5 as the base encoder-decoder models for FiD.

\paragraph{Experimental Details}
For better comparability, we use the same hyperparameter settings for all runs: training is run for 3 epochs, with a batch size of 32 and an initial learning rate of 5e-5. The experiments were conducted on A6000 and A100 NVIDIA GPUs. 

For ($Enc. + Dec.$) models, LLaMA-2 and long sequence techniques, we provide every (input, output) pair in text-to-text format. However, when training $Dec.$ models using the same setting (except for LLaMA-2), we observe poor performance on classification and relation extraction - they either produce malformed labels or just generate continuations for the input text instead of generating the output label. While we did not investigate this deeper, this indicates that long inputs might be particularly problematic for $Dec.$ models. 
Based on these observations, we follow a slightly different setup for $Dec.$ models (except LLaMA-2) on these tasks: the final prediction is made by first encoding the input, then applying a classification head to the last token. For multi-label classification, we continue to follow the text-to-text format. All $Enc. + Dec.$ and $Dec.$ models have an input length of 1024 tokens, while LongT5 and FiD are evaluated with three different token lengths: 2048, 3072, and 4096.

\paragraph{Metrics} 
For all classification and relation extraction tasks in \dataset{}, we report performance using the \textit{Accuracy} metric. However, for RFHD 2014, which is the only dataset for multi-label classification, we use the $F_1$-score metric. For LLaMA-2, we report \textit{lenient accuracy} for all tasks, which post-processes predictions to exclude any unnecessary text generated aside from the predicted label to determine the final \textit{accuracy}.

\subsection{Results}
\input{tables/long_doc_results}

Table \ref{tab:baseline_results} presents the performance of all general domain and medical domain LLMs (denoted as baseline models) that were benchmarked on \dataset, while Table \ref{tab:long_doc_results} showcases the performance of the two long sequence handling techniques we tested.

\paragraph{Baseline Models}
From Table \ref{tab:baseline_results}, we can observe that overall, the average performance of all benchmarked models on \dataset{} is low ($\sim 52\%$). Among $Enc. + Dec.$ models, medical LLMs generally outperform the general domain model (FLAN-T5) on most datasets (five out of seven). Additionally, both medical LLMs, In-BoXBART, Clinical-T5 and SciFive, are competitive with each other. For $Dec.$ only models, we see the reverse - LLaMA-2 actually outperforms both medical LLMs on most datasets (five out of seven). Note that due to the difference in how both model families (except LLaMA-2) were evaluated on classification and relation extraction tasks, $Enc. + Dec.$ models and $Dec.$ only model results cannot be directly compared on most datasets except for RFHD 2014, on which we see that $Enc. + Dec.$ models prevail. Furthermore, we observe that all models consistently exhibit lower scores on datasets with higher input lengths such as ADE 2018 and Cohort Selection (refer to Table \ref{tab:stat_dataset} for length comparison), indicating that incorporating longer input contexts could potentially help. Lastly, we observe that as the model size increases, its capability in handling longer texts improves. This enhancement is evident in the improved results across five out of seven datasets in Table \ref{tab:baseline_results} for LLaMA-2 (7B), compared to other models (<2.7B).

\paragraph{Long Sequence Techniques}
From Table \ref{tab:long_doc_results}, we can see that adding more input context provides mixed results - improving performance over baseline models on some datasets but not others. We also observe that performance on many datasets continues to increase with increasing input token length (from 2048 to 4096 token length). To further analyze the mixed performance of long sequence techniques, we conduct a qualitative error analysis.

\paragraph{Using Clinical vs Biomedical Base Models for FiD}
From Table \ref{tab:fid_clinical_results}, we can see that adding clinical weights shows marginal performance improvement over using FiD with a biomedical base model on many datasets but not all with the largest input token length (4096). We also observe that FiD (w/ClinicalT5) shows mixed performance on many datasets for 2048 and 3072 input token lengths.

\input{tables/fid_clinical}

\subsection{Qualitative Analysis}
\paragraph{Why is \dataset{} difficult for long document models?}
We first perform a qualitative error analysis on one of the datasets on which long document models provide no performance improvement over baseline models: cohort selection. We randomly sample 50 cases in which both techniques get wrong and observe three broad categories of errors. The first category is caused due to very few occurrences and/or late occurrences (i.e., outside our maximum context length of 4096 tokens) of informative cues necessary for the task. The second category occurs because long document techniques, despite increased context length, still do not possess an awareness of EHR document structure (e.g., family history typically does not contain conditions present in the patient) or the ability to deal with longitudinal records (e.g., later test results should override earlier ones). The third category of errors is caused not due to input length, but rather consists of false positives caused by the presence of comorbidities, similar symptoms, variants of the target prediction, etc., which require precise clinical inference to resolve.

\paragraph{Why do models lag behind human performance?} 
Despite mixed overall results, long document techniques do provide strong performance improvements on some datasets, but they still lag behind human performance. We perform an analysis of 50 randomly sampled error cases from one of these datasets (obesity 2008) and observe the same three categories of errors, with $\sim$80\% errors falling into the third category (requiring precise clinical inference).

These observations indicate two potential avenues to further improve model performance on \dataset{}: (i) exploring techniques that perform \emph{relevant sentence selection} in addition to using increased context length, and (ii) developing pretraining and finetuning techniques that equip models with the ability to handle document structure and longitudinality. 

%% file: tables/long_doc_results.tex
\begin{table*}
\setlength\tabcolsep{4.0pt}
\setlength{\belowcaptionskip}{-10pt}
\centering
\footnotesize
\resizebox{\linewidth}{!}{
\begin{tabular}{c|ccc|ccc}
\toprule
                                     & \multicolumn{3}{c|}{\textbf{LongT5}}                                  & \multicolumn{3}{c}{\textbf{FiD (w/SCIFIVE)}}                                     \\ \cmidrule{2-7} 
                                     
\multirow{-2}{*}{\textbf{Dataset}}            & \multicolumn{1}{c|}{\textbf{2048}} & \multicolumn{1}{c|}{\textbf{3072}} & \textbf{4096} & \multicolumn{1}{c|}{\textbf{2048}} & \multicolumn{1}{c|}{\textbf{3072}} & \textbf{4096} \\ \midrule

Smoking 2006                         & \multicolumn{1}{c|}{53.85\tcbox[colback=light_red]{\footnotesize{10.6\% $\downarrow$}}  }    & \multicolumn{1}{c|}{58.65\tcbox[colback=light_red]{\footnotesize{5.76\% $\downarrow$}}}    & 55.77\tcbox[colback=light_red]{\footnotesize{8.64\% $\downarrow$}}    & \multicolumn{1}{c|}{60.26\tcbox[colback=light_red]{\footnotesize{4.15\% $\downarrow$}}}    & \multicolumn{1}{c|}{62.03\tcbox[colback=light_red]{\footnotesize{2.38\% $\downarrow$}}}    & 64.42\tcbox{\footnotesize{0.01\% $\uparrow$}}    \\

Obesity 2008 & \multicolumn{1}{c|}{71.87\tcbox{\footnotesize{0.01\% $\uparrow$}}}    & \multicolumn{1}{c|}{76.79\tcbox{\footnotesize{4.93\% $\uparrow$}}}    & 77.73\tcbox{\footnotesize{5.87\% $\uparrow$}}    & \multicolumn{1}{c|}{64.82\tcbox[colback=light_red]{\footnotesize{7.04\% $\downarrow$}}}    & \multicolumn{1}{c|}{71.32\tcbox[colback=light_red]{\footnotesize{0.54\% $\downarrow$}}}    & 73.15\tcbox{\footnotesize{1.27\% $\uparrow$}}    \\


Assertions 2010                        & \multicolumn{1}{c|}{67.85\tcbox[colback=light_red]{\footnotesize{1.01\% $\downarrow$}}}    & \multicolumn{1}{c|}{68.07\tcbox[colback=light_red]{\footnotesize{0.79\% $\downarrow$}}}    & 67.76\tcbox[colback=light_red]{\footnotesize{1.10\% $\downarrow$}}    & \multicolumn{1}{c|}{67.14\tcbox[colback=light_red]{\footnotesize{1.72\% $\downarrow$}}}    & \multicolumn{1}{c|}{66.95\tcbox[colback=light_red]{\footnotesize{1.91\% $\downarrow$}}}    & 66.71\tcbox[colback=light_red]{\footnotesize{2.15\% $\downarrow$}}    \\ 

Temporal RE                   & \multicolumn{1}{c|}{60.73\tcbox{\footnotesize{4.20\% $\uparrow$}}}    & \multicolumn{1}{c|}{57.96\tcbox{\footnotesize{1.43\% $\uparrow$}}}    & 72.89\tcbox{\footnotesize{16.4\% $\uparrow$}}    & \multicolumn{1}{c|}{58.81\tcbox{\footnotesize{2.28\% $\uparrow$}}}    & \multicolumn{1}{c|}{60.17\tcbox{\footnotesize{3.64\% $\uparrow$}}}    & 63.21\tcbox{\footnotesize{6.68\% $\uparrow$}}    \\

RFHD 2014                            & \multicolumn{1}{c|}{45.07\tcbox[colback=light_red]{\footnotesize{21.6\% $\downarrow$}}}    & \multicolumn{1}{c|}{45.32\tcbox[colback=light_red]{\footnotesize{21.3\% $\downarrow$}}}    & 44.44\tcbox[colback=light_red]{\footnotesize{22.2\% $\downarrow$}}    & \multicolumn{1}{c|}{70.65\tcbox{\footnotesize{4.01\% $\uparrow$}}}    & \multicolumn{1}{c|}{76.16\tcbox{\footnotesize{10.1\% $\uparrow$}}}    & 78.60\tcbox{\footnotesize{12.0\% $\uparrow$}}    \\ 

Cohort Selection                     & \multicolumn{1}{c|}{56.35\tcbox{\footnotesize{8.68\% $\uparrow$}}}    & \multicolumn{1}{c|}{57.70\tcbox{\footnotesize{10.1\% $\uparrow$}}}    & 48.30\tcbox{\footnotesize{0.63\% $\uparrow$}}   & \multicolumn{1}{c|}{48.66\tcbox{\footnotesize{0.99\% $\uparrow$}}}    & \multicolumn{1}{c|}{46.87\tcbox[colback=light_red]{\footnotesize{0.80\% $\downarrow$}}}    & 44.28\tcbox[colback=light_red]{\footnotesize{3.39\% $\downarrow$}}    \\

ADE 2018                             & \multicolumn{1}{c|}{18.12\tcbox[colback=light_red]{\footnotesize{1.10\% $\downarrow$}}}    & \multicolumn{1}{c|}{17.83\tcbox[colback=light_red]{\footnotesize{1.39\% $\downarrow$}}}    & 46.58\tcbox{\footnotesize{27.4\% $\uparrow$}}    & \multicolumn{1}{c|}{17.58\tcbox[colback=light_red]{\footnotesize{1.64\% $\downarrow$}}}    & \multicolumn{1}{c|}{29.15\tcbox{\footnotesize{9.93\% $\uparrow$}}}    & 46.94\tcbox{\footnotesize{27.7\% $\uparrow$}}    \\ \bottomrule
\end{tabular}
}
\caption{Performance of long document techniques, LongT5 and FiD, on \dataset. All results are presented in \%. Green indicates improvement and red indicates degradation in performance compared to the best performing $Enc. + Dec.$ model.}
\label{tab:long_doc_results}
\end{table*}

%% file: tables/fid_clinical.tex
\begin{table}[]
\setlength\tabcolsep{4.0pt}
\setlength{\belowcaptionskip}{-10pt}
\centering
\footnotesize
\resizebox{\linewidth}{!}{
\begin{tabular}{c|ccc}
\toprule
                                         & \multicolumn{3}{c}{\textbf{FiD (w/ClinicalT5)}}                                        \\ \cmidrule{2-4} 
\multirow{-2}{*}{\textbf{Dataset}}       & \multicolumn{1}{c|}{\textbf{2048}} & \multicolumn{1}{c|}{\textbf{3072}} & \textbf{4096} \\ \midrule
Smoking 2006     & \multicolumn{1}{c|}{56.73\tcbox[colback=light_red]{\footnotesize{3.53\% $\downarrow$}}}         & \multicolumn{1}{c|}{60.58\tcbox[colback=light_red]{\footnotesize{1.45\% $\downarrow$}}}         & \multicolumn{1}{c}{60.58\tcbox[colback=light_red]{\footnotesize{3.84\% $\downarrow$}}}         \\ 
Obesity 2008     & \multicolumn{1}{c|}{64.36\tcbox[colback=light_red]{\footnotesize{0.46\% $\downarrow$}}}         & \multicolumn{1}{c|}{73.00\tcbox{\footnotesize{1.68\% $\uparrow$}}}         & \multicolumn{1}{c}{74.20\tcbox{\footnotesize{1.05\% $\uparrow$}}}         \\ 
Assertions 2010  & \multicolumn{1}{c|}{66.71\tcbox[colback=light_red]{\footnotesize{0.43\% $\downarrow$}}}         & \multicolumn{1}{c|}{66.92\tcbox[colback=light_red]{\footnotesize{0.03\% $\downarrow$}}}         & \multicolumn{1}{c}{67.06\tcbox{\footnotesize{0.35\% $\uparrow$}}}         \\ 
Temporal RE      & \multicolumn{1}{c|}{58.37\tcbox[colback=light_red]{\footnotesize{0.44\% $\downarrow$}}}         & \multicolumn{1}{c|}{60.53\tcbox{\footnotesize{0.36\% $\uparrow$}}}         & \multicolumn{1}{c}{63.79\tcbox{\footnotesize{0.58\% $\uparrow$}}}         \\ 
RFHD 2014        & \multicolumn{1}{c|}{60.65\tcbox[colback=light_red]{\footnotesize{10.0\% $\downarrow$}}}         & \multicolumn{1}{c|}{65.46\tcbox[colback=light_red]{\footnotesize{10.7\% $\downarrow$}}}         & \multicolumn{1}{c}{68.76\tcbox[colback=light_red]{\footnotesize{9.84\% $\downarrow$}}}         \\ 
Cohort Selection & \multicolumn{1}{c|}{48.12\tcbox[colback=light_red]{\footnotesize{0.54\% $\downarrow$}}}         & \multicolumn{1}{c|}{46.51\tcbox[colback=light_red]{\footnotesize{0.36\% $\downarrow$}}}         & \multicolumn{1}{c}{46.33\tcbox{\footnotesize{2.23\% $\uparrow$}}}         \\ 
ADE 2018         & \multicolumn{1}{c|}{17.73\tcbox{\footnotesize{0.15\% $\uparrow$}}}         & \multicolumn{1}{c|}{29.36\tcbox{\footnotesize{0.21\% $\uparrow$}}}         & \multicolumn{1}{c}{47.07\tcbox{\footnotesize{0.13\% $\uparrow$}}}         \\ \bottomrule
\end{tabular}
}
\caption{Performance of long document technique FiD (w/ClinicalT5) on \dataset. All results are presented in \%. Green indicates improvement and red indicates degradation in performance compared to the FiD (w/SCIFIVE) from Table \ref{tab:long_doc_results}.}
\label{tab:fid_clinical_results}
\end{table}

%% file: 4_related_work.tex
\section{Related Work}
\label{sec:related_work}

Prior work in the general domain has developed benchmarks to evaluate the ability of transformer-based models to handle long sequence tasks \citep{tay2021long, shaham-etal-2022-scrolls}. These benchmarks motivated the design of several techniques capable of handling long input sequences (see \citet{dong2023survey,tay2022efficient,fournier2021practical} for detailed surveys), which can broadly be divided into two categories: (i) architecture-focused approaches (e.g., developing sparse or hierarchical attention mechanisms), and (ii) data-focused approaches (e.g., chunking or sub-selecting input). However, most of these methods have not been systematically and broadly tested in the clinical domain due to the lack of a comprehensive benchmark which we try to address.

In the clinical domain, some prior work has explored architecture-focused long document approaches \cite{si2021three, li2022clinical, cahyawijaya-etal-2022-long}, however, their evaluation is limited to a handful of tasks. \dataset{}, on the other hand, covers a broad range of tasks and datasets in the clinical domain with longer input token lengths ($>5k$ in many cases) for more comprehensive and systematic evaluation.


%% file: 5_conclusion.tex
\section{Conclusions}

We introduced \dataset, a collection of seven carefully curated clinical datasets, aimed to comprehensively and systematically investigate performance of clinical language models on long text sequences. \dataset{} covers three task types: text classification, relation extraction and multi-label classification and various input types such as longitudinal records and discharge summaries. We benchmark the performance of eight general and medical domains LLMs on \dataset{}, and show that they do not achieve good performance. Furthermore, we also investigate two techniques to handle long sequences and our results reveal that though these methods sometimes improve performance, showing the importance of long input context, there is substantial room for improvement. We believe that \dataset{} can serve as an important benchmark for developing long sequence techniques tailored to the clinical domain.

\section*{Limitations}

In this work, we've largely benchmarked state-of-the-art general and biomedical language models on \dataset, but we also plan to benchmark more clinical language models. Currently, \dataset{} is limited in terms of task variety since it primarily consists of different types of classification tasks. This is largely because it is challenging to find shareable datasets across various task types in the clinical domain, but we plan to further increase task variety in this benchmark. Additionally, we hope to expand our analysis to include the most recent large language models such as GPT-4 and ChatGPT on \dataset. Our observation that existing long document models still struggle on \dataset{}, also suggests that it may be interesting to conduct detailed analysis of different aspects such as model understanding of clinical document structure and better clinical tokenization, which we have left to future work.

%% file: emnlp2023.bbl
\begin{thebibliography}{30}
\expandafter\ifx\csname natexlab\endcsname\relax\def\natexlab#1{#1}\fi

\bibitem[{Black et~al.(2022)Black, Biderman, Hallahan, Anthony, Gao, Golding,
  He, Leahy, McDonell, Phang, Pieler, Prashanth, Purohit, Reynolds, Tow, Wang,
  and Weinbach}]{black-etal-2022-gpt}
Sidney Black, Stella Biderman, Eric Hallahan, Quentin Anthony, Leo Gao,
  Laurence Golding, Horace He, Connor Leahy, Kyle McDonell, Jason Phang,
  Michael Pieler, Usvsn~Sai Prashanth, Shivanshu Purohit, Laria Reynolds,
  Jonathan Tow, Ben Wang, and Samuel Weinbach. 2022.
\newblock \href {https://doi.org/10.18653/v1/2022.bigscience-1.9}
  {{GPT}-{N}eo{X}-20{B}: An open-source autoregressive language model}.
\newblock In \emph{Proceedings of BigScience Episode {\#}5 -- Workshop on
  Challenges {\&} Perspectives in Creating Large Language Models}, pages
  95--136, virtual+Dublin. Association for Computational Linguistics.

\bibitem[{Cahyawijaya et~al.(2022)Cahyawijaya, Wilie, Lovenia, Zhong, Zhong,
  Ip, and Fung}]{cahyawijaya-etal-2022-long}
Samuel Cahyawijaya, Bryan Wilie, Holy Lovenia, Huan Zhong, MingQian Zhong,
  Yuk-Yu~Nancy Ip, and Pascale Fung. 2022.
\newblock \href {https://aclanthology.org/2022.louhi-1.19} {How long is enough?
  exploring the optimal intervals of long-range clinical note language
  modeling}.
\newblock In \emph{Proceedings of the 13th International Workshop on Health
  Text Mining and Information Analysis (LOUHI)}, pages 160--172, Abu Dhabi,
  United Arab Emirates (Hybrid). Association for Computational Linguistics.

\bibitem[{Chung et~al.(2022)Chung, Hou, Longpre, Zoph, Tay, Fedus, Li, Wang,
  Dehghani, Brahma et~al.}]{chung2022scaling}
Hyung~Won Chung, Le~Hou, Shayne Longpre, Barret Zoph, Yi~Tay, William Fedus,
  Eric Li, Xuezhi Wang, Mostafa Dehghani, Siddhartha Brahma, et~al. 2022.
\newblock Scaling instruction-finetuned language models.
\newblock \emph{arXiv preprint arXiv:2210.11416}.

\bibitem[{Dong et~al.(2023)Dong, Tang, Li, and Zhao}]{dong2023survey}
Zican Dong, Tianyi Tang, Lunyi Li, and Wayne~Xin Zhao. 2023.
\newblock A survey on long text modeling with transformers.
\newblock \emph{arXiv preprint arXiv:2302.14502}.

\bibitem[{Fournier et~al.(2021)Fournier, Caron, and
  Aloise}]{fournier2021practical}
Quentin Fournier, Ga{\'e}tan~Marceau Caron, and Daniel Aloise. 2021.
\newblock A practical survey on faster and lighter transformers.
\newblock \emph{ACM Computing Surveys}.

\bibitem[{Guo et~al.(2022)Guo, Ainslie, Uthus, Ontanon, Ni, Sung, and
  Yang}]{guo-etal-2022-longt5}
Mandy Guo, Joshua Ainslie, David Uthus, Santiago Ontanon, Jianmo Ni, Yun-Hsuan
  Sung, and Yinfei Yang. 2022.
\newblock \href {https://doi.org/10.18653/v1/2022.findings-naacl.55}
  {{L}ong{T}5: {E}fficient text-to-text transformer for long sequences}.
\newblock In \emph{Findings of the Association for Computational Linguistics:
  NAACL 2022}, pages 724--736, Seattle, United States. Association for
  Computational Linguistics.

\bibitem[{Henry et~al.(2020)Henry, Buchan, Filannino, Stubbs, and
  Uzuner}]{henry20202018}
Sam Henry, Kevin Buchan, Michele Filannino, Amber Stubbs, and Ozlem Uzuner.
  2020.
\newblock 2018 n2c2 shared task on adverse drug events and medication
  extraction in electronic health records.
\newblock \emph{Journal of the American Medical Informatics Association},
  27(1):3--12.

\bibitem[{Izacard and Grave(2021)}]{izacard-grave-2021-leveraging}
Gautier Izacard and Edouard Grave. 2021.
\newblock \href {https://doi.org/10.18653/v1/2021.eacl-main.74} {Leveraging
  passage retrieval with generative models for open domain question answering}.
\newblock In \emph{Proceedings of the 16th Conference of the European Chapter
  of the Association for Computational Linguistics: Main Volume}, pages
  874--880, Online. Association for Computational Linguistics.

\bibitem[{Kind and Smith(2008)}]{kind2008documentation}
Amy~JH Kind and Maureen~A Smith. 2008.
\newblock Documentation of mandated discharge summary components in transitions
  from acute to subacute care.
\newblock \emph{Advances in patient safety: new directions and alternative
  approaches (Vol. 2: culture and redesign)}.

\bibitem[{Kumar et~al.(2015)Kumar, Stubbs, Shaw, and
  Uzuner}]{kumar2015creation}
Vishesh Kumar, Amber Stubbs, Stanley Shaw, and {\"O}zlem Uzuner. 2015.
\newblock Creation of a new longitudinal corpus of clinical narratives.
\newblock \emph{Journal of biomedical informatics}, 58:S6--S10.

\bibitem[{Lewis et~al.(2020)Lewis, Ott, Du, and
  Stoyanov}]{lewis-etal-2020-pretrained}
Patrick Lewis, Myle Ott, Jingfei Du, and Veselin Stoyanov. 2020.
\newblock \href {https://doi.org/10.18653/v1/2020.clinicalnlp-1.17} {Pretrained
  language models for biomedical and clinical tasks: Understanding and
  extending the state-of-the-art}.
\newblock In \emph{Proceedings of the 3rd Clinical Natural Language Processing
  Workshop}, pages 146--157, Online. Association for Computational Linguistics.

\bibitem[{Li et~al.(2022)Li, Wehbe, Ahmad, Wang, and Luo}]{li2022clinical}
Yikuan Li, Ramsey~M Wehbe, Faraz~S Ahmad, Hanyin Wang, and Yuan Luo. 2022.
\newblock Clinical-longformer and clinical-bigbird: Transformers for long
  clinical sequences.
\newblock \emph{arXiv preprint arXiv:2201.11838}.

\bibitem[{Lu et~al.(2022)Lu, Dou, and Nguyen}]{lu-etal-2022-clinicalt5}
Qiuhao Lu, Dejing Dou, and Thien Nguyen. 2022.
\newblock \href {https://doi.org/10.18653/v1/2022.findings-emnlp.398}
  {{C}linical{T}5: A generative language model for clinical text}.
\newblock In \emph{Findings of the Association for Computational Linguistics:
  EMNLP 2022}, pages 5436--5443, Abu Dhabi, United Arab Emirates. Association
  for Computational Linguistics.

\bibitem[{Luo et~al.(2022)Luo, Sun, Xia, Qin, Zhang, Poon, and
  Liu}]{luo2022biogpt}
Renqian Luo, Liai Sun, Yingce Xia, Tao Qin, Sheng Zhang, Hoifung Poon, and
  Tie-Yan Liu. 2022.
\newblock Biogpt: generative pre-trained transformer for biomedical text
  generation and mining.
\newblock \emph{Briefings in Bioinformatics}, 23(6).

\bibitem[{Parmar et~al.(2022)Parmar, Mishra, Purohit, Luo, Mohammad, and
  Baral}]{parmar-etal-2022-boxbart}
Mihir Parmar, Swaroop Mishra, Mirali Purohit, Man Luo, Murad Mohammad, and
  Chitta Baral. 2022.
\newblock \href {https://doi.org/10.18653/v1/2022.findings-naacl.10}
  {In-{B}o{XBART}: Get instructions into biomedical multi-task learning}.
\newblock In \emph{Findings of the Association for Computational Linguistics:
  NAACL 2022}, pages 112--128, Seattle, United States. Association for
  Computational Linguistics.

\bibitem[{Peng et~al.(2023)Peng, Yang, Chen, Smith, PourNejatian, Costa,
  Martin, Flores, Zhang, Magoc et~al.}]{peng2023study}
Cheng Peng, Xi~Yang, Aokun Chen, Kaleb~E Smith, Nima PourNejatian, Anthony~B
  Costa, Cheryl Martin, Mona~G Flores, Ying Zhang, Tanja Magoc, et~al. 2023.
\newblock A study of generative large language model for medical research and
  healthcare.
\newblock \emph{arXiv preprint arXiv:2305.13523}.

\bibitem[{Phan et~al.(2021)Phan, Anibal, Tran, Chanana, Bahadroglu, Peltekian,
  and Altan-Bonnet}]{phan2021scifive}
Long~N Phan, James~T Anibal, Hieu Tran, Shaurya Chanana, Erol Bahadroglu, Alec
  Peltekian, and Gr{\'e}goire Altan-Bonnet. 2021.
\newblock Scifive: a text-to-text transformer model for biomedical literature.
\newblock \emph{arXiv preprint arXiv:2106.03598}.

\bibitem[{Raffel et~al.(2020)Raffel, Shazeer, Roberts, Lee, Narang, Matena,
  Zhou, Li, and Liu}]{raffel2020exploring}
Colin Raffel, Noam Shazeer, Adam Roberts, Katherine Lee, Sharan Narang, Michael
  Matena, Yanqi Zhou, Wei Li, and Peter~J Liu. 2020.
\newblock Exploring the limits of transfer learning with a unified text-to-text
  transformer.
\newblock \emph{The Journal of Machine Learning Research}, 21(1):5485--5551.

\bibitem[{Shaham et~al.(2022)Shaham, Segal, Ivgi, Efrat, Yoran, Haviv, Gupta,
  Xiong, Geva, Berant, and Levy}]{shaham-etal-2022-scrolls}
Uri Shaham, Elad Segal, Maor Ivgi, Avia Efrat, Ori Yoran, Adi Haviv, Ankit
  Gupta, Wenhan Xiong, Mor Geva, Jonathan Berant, and Omer Levy. 2022.
\newblock \href {https://aclanthology.org/2022.emnlp-main.823} {{SCROLLS}:
  Standardized {C}ompa{R}ison over long language sequences}.
\newblock In \emph{Proceedings of the 2022 Conference on Empirical Methods in
  Natural Language Processing}, pages 12007--12021, Abu Dhabi, United Arab
  Emirates. Association for Computational Linguistics.

\bibitem[{Si and Roberts(2021)}]{si2021three}
Yuqi Si and Kirk Roberts. 2021.
\newblock Three-level hierarchical transformer networks for long-sequence and
  multiple clinical documents classification.
\newblock \emph{arXiv preprint arXiv:2104.08444}.

\bibitem[{Singhal et~al.(2022)Singhal, Azizi, Tu, Mahdavi, Wei, Chung, Scales,
  Tanwani, Cole-Lewis, Pfohl et~al.}]{singhal2022large}
Karan Singhal, Shekoofeh Azizi, Tao Tu, S~Sara Mahdavi, Jason Wei, Hyung~Won
  Chung, Nathan Scales, Ajay Tanwani, Heather Cole-Lewis, Stephen Pfohl, et~al.
  2022.
\newblock Large language models encode clinical knowledge.
\newblock \emph{arXiv preprint arXiv:2212.13138}.

\bibitem[{Stubbs et~al.(2019)Stubbs, Filannino, Soysal, Henry, and
  Uzuner}]{stubbs2019cohort}
Amber Stubbs, Michele Filannino, Ergin Soysal, Samuel Henry, and {\"O}zlem
  Uzuner. 2019.
\newblock Cohort selection for clinical trials: n2c2 2018 shared task track 1.
\newblock \emph{Journal of the American Medical Informatics Association},
  26(11):1163--1171.

\bibitem[{Sun et~al.(2013)Sun, Rumshisky, and Uzuner}]{sun2013evaluating}
Weiyi Sun, Anna Rumshisky, and Ozlem Uzuner. 2013.
\newblock Evaluating temporal relations in clinical text: 2012 i2b2 challenge.
\newblock \emph{Journal of the American Medical Informatics Association},
  20(5):806--813.

\bibitem[{Tay et~al.(2021)Tay, Dehghani, Abnar, Shen, Bahri, Pham, Rao, Yang,
  Ruder, and Metzler}]{tay2021long}
Yi~Tay, Mostafa Dehghani, Samira Abnar, Yikang Shen, Dara Bahri, Philip Pham,
  Jinfeng Rao, Liu Yang, Sebastian Ruder, and Donald Metzler. 2021.
\newblock \href {https://openreview.net/forum?id=qVyeW-grC2k} {Long range
  arena: A benchmark for efficient transformers}.
\newblock In \emph{International Conference on Learning Representations}.

\bibitem[{Tay et~al.(2022)Tay, Dehghani, Bahri, and Metzler}]{tay2022efficient}
Yi~Tay, Mostafa Dehghani, Dara Bahri, and Donald Metzler. 2022.
\newblock Efficient transformers: A survey.
\newblock \emph{ACM Computing Surveys}, 55(6):1--28.

\bibitem[{Touvron et~al.(2023)Touvron, Martin, Stone, Albert, Almahairi,
  Babaei, Bashlykov, Batra, Bhargava, Bhosale et~al.}]{touvron2023llama}
Hugo Touvron, Louis Martin, Kevin Stone, Peter Albert, Amjad Almahairi, Yasmine
  Babaei, Nikolay Bashlykov, Soumya Batra, Prajjwal Bhargava, Shruti Bhosale,
  et~al. 2023.
\newblock Llama 2: Open foundation and fine-tuned chat models.
\newblock \emph{arXiv preprint arXiv:2307.09288}.

\bibitem[{Uzuner(2009)}]{uzuner2009recognizing}
{\"O}zlem Uzuner. 2009.
\newblock Recognizing obesity and comorbidities in sparse data.
\newblock \emph{Journal of the American Medical Informatics Association},
  16(4):561--570.

\bibitem[{Uzuner et~al.(2008)Uzuner, Goldstein, Luo, and
  Kohane}]{uzuner2008identifying}
{\"O}zlem Uzuner, Ira Goldstein, Yuan Luo, and Isaac Kohane. 2008.
\newblock Identifying patient smoking status from medical discharge records.
\newblock \emph{Journal of the American Medical Informatics Association},
  15(1):14--24.

\bibitem[{Uzuner et~al.(2011)Uzuner, South, Shen, and DuVall}]{uzuner20112010}
{\"O}zlem Uzuner, Brett~R South, Shuying Shen, and Scott~L DuVall. 2011.
\newblock 2010 i2b2/va challenge on concepts, assertions, and relations in
  clinical text.
\newblock \emph{Journal of the American Medical Informatics Association},
  18(5):552--556.

\bibitem[{Venigalla et~al.(2022)Venigalla, Frankle, and
  Carbin}]{venigalla2022biomedlm}
A~Venigalla, J~Frankle, and M~Carbin. 2022.
\newblock Biomedlm: a domain-specific large language model for biomedical text.
\newblock \emph{MosaicML. Accessed: Dec}, 23:3.

\end{thebibliography}
